\pdfoutput=1
\PassOptionsToPackage{hyphens}{url}
\documentclass[11pt]{article}

\usepackage{dcolumn}

\usepackage{acl}

\usepackage{times}
\usepackage{latexsym}
\usepackage{tablefootnote}
\usepackage{amsmath}
\usepackage{tikz}
\usepackage[T1]{fontenc}

\usepackage[utf8]{inputenc}

\usepackage{microtype}

\def\checkmark{\tikz\fill[scale=0.4](0,.35) -- (.25,0) -- (1,.7) -- (.25,.15) -- cycle;} 

\title{How to Find Strong Summary Coherence Measures? A Toolbox and a Comparative Study for Summary Coherence Measure Evaluation}

\author{Julius Steen \enspace Katja Markert\\
  Department of Computational Linguistics \\
  Heidelberg University\\
  69120 Heidelberg, Germany \\
  {\tt (steen|markert)@cl.uni-heidelberg.de}}
\date{}

\begin{document}
\maketitle
\begin{abstract}
Automatically evaluating the coherence of summaries is of great significance both to enable cost-efficient summarizer evaluation and as a tool for improving coherence by selecting high-scoring candidate summaries. While many different approaches have been suggested to model summary coherence, they are often evaluated using disparate datasets and metrics. This makes it difficult to understand their relative performance and identify ways forward towards better summary coherence modelling.
In this work, we conduct a large-scale investigation of various methods for summary coherence modelling on an even playing field. Additionally, we introduce two novel analysis measures, \textit{intra-system correlation} and \textit{bias matrices}, that help identify biases in coherence measures and provide robustness against system-level confounders. While none of the currently available automatic coherence
measures are able to assign reliable coherence scores to system summaries across all evaluation metrics, 
large-scale language models fine-tuned on self-supervised tasks show promising results, as long as fine-tuning takes into account that they need to generalize across different summary lengths.
\end{abstract}

\section{Introduction}

Automatically generated summaries should not only be informative, but also well-written and coherent. While informativeness is routinely evaluated automatically with ROUGE \cite{lin_rouge_2004}, there is no agreement on how to evaluate summary coherence. However, automatic evaluation is highly desirable to reduce evaluation costs and as a tool for improving summarizer output, e.g. as reranker.

Many coherence measures \textbf{(CMs)} have been suggested for automatically assigning a coherence score to a summary, including learning from human judgements \citep{barzilay-lapata-2008-modeling, tien-nguyen-joty-2017-neural, xenouleas-etal-2019-sum, mesgar-etal-2021-neural-graph}, learning from the \textit{shuffle task}  \citep{mohiuddin_rethinking_2021, jwalapuram-etal-2022-rethinking}, where models are trained to discriminate original documents from documents with randomized sentence order \citep{barzilay-lapata-2008-modeling}, using next sentence prediction as a proxy task \citep{koto2022ffci}, and finally unsupervised measures that exploit heuristics \citep{pitler-etal-2010-automatic, zhu-bhat-2020-gruen} or 
large-scale LMs \citep{yuan_bartscore_2021}. CM performance is then evaluated by comparing the automatic scores to human coherence scores on a set of system summaries.

However, this evaluation is often conducted on disparate datasets. It also often uses system outputs from
DUC conferences \citep{barzilay-lapata-2008-modeling, tien-nguyen-joty-2017-neural, xenouleas-etal-2019-sum, mesgar-etal-2021-neural-graph}, which do not necessarily represent recent advances in text summarizers.
In addition, there is no agreement on \textit{how} the CM scores should be compared to human scores. 
System-level correlation
\citep{xenouleas-etal-2019-sum, fabbri_summeval_2021}, pairwise ranking accuracy \citep{barzilay-lapata-2008-modeling, tien-nguyen-joty-2017-neural, mesgar-etal-2021-neural-graph} and summary-level correlation \citep{yuan_bartscore_2021} have all been suggested as evaluation metrics \textbf{(EMs)}.

This makes it hard to ascertain the state of summary coherence modelling and to identify promising directions for future research. %
We attack this problem by making the following contributions:

\begin{itemize}
    \item We show that current EMs provide an incomplete picture of CM performance as they focus on comparing summaries generated by different summarizers, which includes many easy decisions due to the large performance gaps between them. Additionally, they are vulnerable to CMs exploiting confounding system properties
    to correctly rank systems without modelling coherence.
    \item We introduce a new EM, \textit{intra-system correlation}, that measures performance within the summaries generated by a single summarizer and is both more challenging and more resilient against system-level confounders.
    \item We introduce \textit{bias matrices} as a novel analysis tool that allow to easily detect when CMs are biased towards specific summarizers.
\end{itemize}

\noindent 
Using these insights, we conduct a large-scale comparison of CMs on the recent \textit{SummEval} dataset \citep{fabbri_summeval_2021}. We find that:

\begin{itemize}
    \item All investigated CMs exhibit significant weaknesses under evaluation regimes other than system-level correlation.
   \item Even relatively strong CMs are biased towards outputs of certain summarizers, which raises concern about their generalizability.
    \item SummEval is not conducive to entity-based modelling, which has been successful on many other coherence tasks \citep{barzilay-lapata-2008-modeling,elsner-charniak-2011-extending, tien-nguyen-joty-2017-neural,mesgar-etal-2021-neural-graph}.
    
    \item While most of the shuffle-based models transfer poorly to summaries, which is in line with prior results by \citet{mohiuddin_rethinking_2021}, the most promising performance is achieved by  fine-tuning a masked language model (MLM) on the shuffle task as a \textit{classifier}. We present evidence that this allows the model to adapt  more easily to comparing documents of different content and lengths, highlighting a possible avenue for future work.
\end{itemize}

Code and data for our experiments are available at \url{https://github.com/julmaxi/summary_coherence_evaluation}.

\section{Related Work} \label{sec:related}

\subsection{Coherence Measures for Summarization}

Automatic coherence assessment for summarization has been studied in a variety of settings.
\citet{barzilay-lapata-2008-modeling} establish summary coherence as an \textit{evaluation} task to assess CMs similarly to other downstream tasks such as essay scoring \cite[among others]{jeon-strube-2020-incremental} and readability assessment \cite[among others]{mesgar_graph-based_2015}. Specifically, \citeauthor{barzilay-lapata-2008-modeling} acquire coherence labels for human and system summaries from DUC 2003\footnote{\url{https://duc.nist.gov}}. 
The same dataset has been used  for evaluating subsequent CMs \citep{tien-nguyen-joty-2017-neural, mesgar-etal-2021-neural-graph}.

As a part of \textit{automatic linguistic quality estimation}, summary coherence is modelled alongside other aspects of text quality such as grammaticality, with the direct goal of aiding automatic summary evaluation. Approaches include regression models learned from human annotations \cite{xenouleas-etal-2019-sum} as well as unsupervised approaches \citep{pitler-etal-2010-automatic, zhu-bhat-2020-gruen, yuan_bartscore_2021}. Datasets used for evaluation include the recent SummEval dataset \citep{fabbri_summeval_2021}, assessor judgements from DUC05-07, and the small-scale manually annotated summaries of newsroom \citep{grusky-etal-2018-newsroom}. In parallel work, \citet{koto2022ffci} introduce a coherence measure based on a next sentence prediction task as part of a wider set of measures for summary evaluation that also include focus, coverage and faithfulness. For evaluation, they introduce a novel small-scale dataset based on outputs from BART \citep{lewis_bart_2020} and the pointer generator model \citep{see_get_2017}.
CMs have also been applied in a related setting to improve summarizer quality by explicitly modelling coherence during the summary optimization process \citep{parveen_generating_2017, sharma_entity-driven_2019}.

Finally, summary coherence is also sometimes modelled using measures that make use of human-written reference summaries \citep{fabbri_summeval_2021, zhao_discoscore_2022}. We do not focus on these CMs in our evaluation since they are fundamentally less flexible than reference-free CMs, especially when used in non-evaluation contexts such as reranking.

We provide a detailed description of the CMs used in our study in Section~\ref{sec:measures}.

\subsection{Meta-Evaluation}

In terms of evaluation studies, \citet{mohiuddin_rethinking_2021} conduct a comparative study of five CMs. Their evaluation is conducted on  10 summaries each from 4 recent summarizers as well as  the DUC03 data. Unlike our study, their investigation only encompasses CMs trained via the shuffle task and includes only a small number of summaries.

In concurrent work on assessing system-level correlation, 
\citet{deutsch-etal-2022-examining} 
propose to modify correlation computation by focusing on difficult system comparisons only and computing measure scores on a larger set of summaries. Their approaches are complimentary to our analysis in that they look at 
informativeness instead of coherence and do not address the shortcomings of system-level correlation in the presence of system level confounders. %
Also concurrently, \citet{durmus-etal-2022-spurious} identify spurious correlates  in faithfulness
datasets, which suggests that our methods might be useful beyond coherence evaluation.

\section{Dataset}

\begin{figure}
    \centering
    \includegraphics[width=0.45\textwidth]{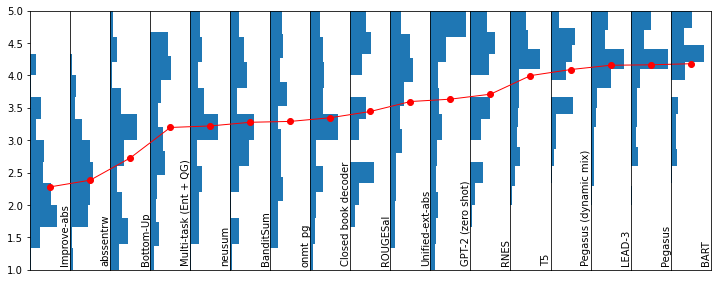}
    \caption{Distribution of human coherence scores for the 17 systems in the SummEval dataset. The red dots indicate the mean score of each system.}
    \label{fig:score_dist}
\end{figure}

We evaluate CMs on the expert annotations in the SummEval dataset\footnote{ \url{https://github.com/Yale-LILY/SummEval}} \cite{fabbri_summeval_2021} which is, to the best of our knowledge, the largest dataset that includes 
such coherence annotations for a variety of state-of-the-art summarizers.
It contains annotations on a 1-5 scale for outputs of 17 systems for 100 documents from the CNN/DM dataset \cite{hermann_teaching_2015} by three annotators each. Figure~\ref{fig:score_dist} highlights two important properties. Firstly, there is a large gap in average performance 
between different summarizers, and secondly, most summarizers exhibit considerable variance in scores.

\section{Evaluating CMs} \label{sec:em_intro}

 CM performance is typically assessed on a set of summaries generated on document set $D$ by a set of summarizers $S$ using the agreement of predicted scores $P = \{P_{(d,s)}| s \in S, d \in D\}$ with human judgements $H = \{H_{(d,s)} | s \in S, d \in D\}$. However, this agreement can be computed in different ways. We identify the following common EMs:

 {\bfseries System-level Correlation $\tau_{sys}$ }  assesses CM performance by correlating
 the \textit{mean} human and \textit{mean} CM scores of the individual summarizers.
 
{\bfseries Pairwise Accuracy $Acc_{pair}$ } assesses CM performance by comparing scores on outputs of two different systems on the same document.

{\bfseries Summary-level Correlation $\tau_{sum}$} compares scores on all generated summaries.

The correlation function used is usually\footnote{While Spearmans $\rho$ is also sometimes used \citep{yuan_bartscore_2021}, the $\tau$ variant we use, $\tau$-b, is more robust to ties, which are common on the five-point rating scale in  SummEval.} Kendall's $\tau$, while the pairwise metric is usually reported as accuracy. However, we can also define the latter in terms of average $\tau$ over all documents. This is equivalent to accuracy when there are no tied scores, with the only difference being the range shift from $[0, 1]$ to $[-1, 1]$.
As both $\tau$ and accuracy are based on pairwise rankings, we can specify all three EMs in terms of the set of pairwise comparisons $C$ they consider, where $C_{sys} \subset 2^{S \times S}$ considers comparisons between averaged system scores and $C_{pair}, C_{sum} \subset 2^{(D \times S) \times (D \times S)}$ consider comparisons between individual summary scores. 

\begin{align*}
C_{sys} = \{(s_i, s_j) | s_i \neq s_j \} \\
C_{pair} = \{ ((d, s_i), (d, s_j)) | s_i \neq s_j \} \\
C_{sum} = \{ ((d_k, s_i), (d_l, s_j)) | (d_k, s_i) \neq (d_l, s_j) \}
\end{align*}

\noindent The EMs pose different demands to CMs: system-level correlation requires a correct ranking of systems according to their average score. Pairwise accuracy requires correct ranking of summaries from different systems but only between summaries produced on the same document. Finally, summary-level correlation requires the correct ranking of any pair of summaries. %

\subsection{A new EM: Intra-System Correlation} \label{sec:sys_lvl_corr}

All three EMs focus on comparisons between summaries generated by different summarizers.
For system-level and pairwise evaluation this arises by construction, whereas for summary-level correlation it is contingent on the dataset structure: On SummEval, less than $6\%$ of comparisons for $\tau_{sum}$ are between summaries of the same summarizer.
We argue that this gives an incomplete view of CM performance for the following reasons:

\begin{enumerate}
   \item SummEval covers summarizers with widely different performance levels (see Figure~\ref{fig:score_dist}), leading current EMs to include many easy decisions. This is unlikely to reflect real-world evaluation of competitive summarizers. 
   
    \item While system-level evaluation is often the primary use case, CMs can also be used in a reranking or ensembling context to select the most coherent summary from a set of candidates. In these situations, summaries are likely to be generated either by a the same summarizer or a set of similarly (high) performing summarization systems. In these cases, system-level EMs offer only limited insight into likely CM performance, since they primarily measure the ability to discriminate between \textit{different} systems with potentially large performance gaps.
    \item EMs might not correlate with coherence per-se but instead with  features that happen to identify good summarizers on a particular dataset. Such \textit{system-level confounders} are unlikely to generalize to new systems and settings. We elaborate on this in Section \ref{sec:confounders}. 
\end{enumerate}

We thus suggest adding a new EM {\bfseries Intra-system Correlation $\tau_{intra}$}, which  we define on comparisons between summaries generated by the same system. This corresponds to considering the following pairs $C_{intra} \subset 2^{(D \times S) \times (D \times S)}$:

\begin{equation*}
    C_{intra} = \{ ((d_k, s), (d_l, s)) | d_k \neq d_l \}
\end{equation*}

It neatly complements pairwise accuracy, as it is essentially the same computation but keeping the summarizer constant instead of the document. Intuitively, this measure both contains far fewer ''easy'' decisions and is much more resilient to any system-level confounders in the data.
We use the average of the intra-system correlation of all systems as the correlation measure.

\subsection{System-level Confounders} \label{sec:confounders}

To assess how EMs behave in the presence of system-level confounders, we investigate two summary features that are unlikely to be generalizable CMs but lead to surprisingly strong correlations:  Capitalization and summarizer architecture.

For capitalization, we count the number of uppercase letters in each summary. This is a purely system-level heuristics, since only three of the 17 summarizers in SummEval produce capital letters\footnote{BART, GPT-2 (zero shot) and Pegasus (dynamic mix)}. For architecture, we assign a score of 1 to each summary from one of the five summarizers that are derived from pretrained transformers in some fashion\footnote{BART, Pegasus, Pegasus dynamic Mix, T5 and GPT-2} and 0 to all others. Neither of the two confounders can, by construction, be a reasonable and generalizable CM.
Additionally, we compute an ''upper-bound'' (UB) that assigns to each summary the mean human score of the system that produced the summary. It simulates perfect system ranking, but no ability to correctly rank summaries within each system.
Since these procedures result in many ties, we also compute a second variant of each confounder where we add small noise to each score. This prevents $\tau$-b from profiting from these ties, while preventing accuracy from unfairly suffering.

\begin{table}
    \centering
    \tiny\begin{tabular}{|l|c|c|c|c|c|c|}
    \hline
     &    Cap. & Cap. (r) & Arch. & Arch. (r) & UB & UB (r) \\\hline
     $\tau_{sys}$        & 0.42  & 0.23  & 0.58 & 0.37 & 1.00 & 1.00 \\
     $\tau_{sum}$      & 0.19  & 0.11  & 0.31 & 0.20 & 0.39 & 0.39 \\
     $\tau_{pair}$   &  0.21    & 0.14  & 0.33    & 0.22 & 0.44 & 0.44  \\
     $Acc_{pair}$   &  0.23    & 0.57  &  0.34   & 0.62 & 0.73 & 0.73  \\
     $\tau_{intra}$   &  -    & -0.03\phantom{-}  &  -   & 0.01 & - & 0.00  \\\hline
    \end{tabular}
    \caption{Results for the confounders and upper bound. $\tau_{intra}$ for the non-random variants is undefined, as
    scores within each system are constant. Scores for the random variants (r) are averaged over 100 runs.}
    \label{tab:shortcuts}
\end{table}

Table~\ref{tab:shortcuts} shows the resulting correlations.  Confounders achieve noticeable correlation with human scores. In particular, system-level correlation comes close to or exceeds the best CM reported originally for SummEval (CHRF \citep{popovic-2017-chrf}, 0.40).
In contrast, using intra-system correlation, the problems of these pseudo-measures become easily apparent.
In practical scenarios, system-level correlation might be a mix of modelling coherence and reliance on confounders. Intra-system  evaluation is an important tool in this context as it is 
robust to system-level confounders.

\section{Coherence Measures} \label{sec:measures}

\begin{table*}
    \centering
    \small\begin{tabular}{|c|c|c|c|c|c|c|c|c|c|}
    \hline
                        & \texttt{EEG}           &    \texttt{EGR}   & \texttt{NEG} & \texttt{UNF} & \texttt{GRA} & \texttt{CCL} & \texttt{SQE} & \texttt{GRU} & \texttt{BAS}            \\\hline
Unsupervised            &  \checkmark{}$^{(a)}$          &  \checkmark{}     &              &              &              &              &              & \checkmark{} & \checkmark{}            \\
Shuffle                 &  &                   & \checkmark{} & \checkmark{} & \checkmark{} & \checkmark{} &              &              & $^{(b)}$    \\
Supervised (DUC03)      &  &                   & \checkmark{} &              & \checkmark{} &              &              &              &                         \\
Supervised (DUC05-07)   &                        &                   &              &              &              &              & \checkmark{} &              &                         \\\hline
    \end{tabular}
    \caption{Training settings for the CMs under investigation. (a) The extended entity grid estimates the multinomial distribution of an entity's role given its prior occurrences. While this needs a dataset to estimate the distribution, it can not be trained as a classifier. (b) BART includes shuffling as a pretraining task. }
    \label{tab:configurations}
\end{table*}

We identify the following families of reference-free CMs for summarization and include representatives of each in our study: \textbf{supervised CMs} trained on human coherence ratings of summaries, \textbf{self-supervised CMs trained on the shuffle task} and \textbf{unsupervised CMs}.
For the supervised setting, we investigate measures trained on data from DUC03  \cite{barzilay-lapata-2008-modeling} as well as  the DUC05-07 dataset used by \citet{xenouleas-etal-2019-sum}. While the DUC03 dataset is set up as a pairwise ranking dataset, the DUC05-07 dataset is used in a regression setting. Table~\ref{tab:configurations} indicates the configurations available for the different CMs.

The \textbf{Extended Entity Grid (\texttt{EEG})} \cite{elsner-charniak-2011-extending} is an extension of the Entity Grid of \citet{barzilay-lapata-2008-modeling}. It represents texts using occurrence patterns of the mentioned entities across sentences. The model uses a generative approach that models the probability of an entity appearing in a specific role in a sentence, given its role in the two preceding sentences.

The \textbf{Entity Graph (\texttt{EGR})} \cite{guinaudeau-strube-2013-graph} constructs a sentence graph of a document by identifying entity overlap between sentences. Two sentences are connected if they share at least one entity, with edge weights decreasing when they are further apart. The score of a document is the average outdegree of sentences, with higher outdegree indicating better coherence. 

The \textbf{Neural Entity Grid (\texttt{NEG})} \cite{tien-nguyen-joty-2017-neural}  applies a convolutional network to the entity grid. The model is trained on a pairwise ranking loss.

The \textbf{Unified Model (\texttt{UNF})} \cite{moon-etal-2019-unified} is a lexical CM that uses a convolutional network to build sentence representations from raw text. The model uses an adapted version of the ranking loss for the shuffle task that is computed only for three sentence windows in which shuffled and original documents differ. We use the model based on ELMo \cite{peters-etal-2018-deep}, as it performs best in the original paper.

The \textbf{Graph-based Neural Coherence Model (\texttt{GRA})} \cite{mesgar-etal-2021-neural-graph} is  a recent CM that combines entity-based representation with lexical information in a graph NN. Like the previous two models, it employs a pairwise ranking loss.

Recently, \citet{laban-etal-2021-transformer} have shown that a RoBERTa-based \cite{liu_roberta_2019} classifier can easily achieve near-perfect results on the shuffling task on WSJ. However, they did not test whether this model can predict coherence on non-artificial tasks. We thus include a simple RoBERTa model that is trained to classify shuffled vs. unshuffled summaries, naming it  {\bfseries Coherence Classifier (\texttt{CCL})}.\footnote{We found that the original WSJ-model does not perform well on SummEval. Thus,  we retrained our own model, using the same RoBERTA checkpoint as a basis.}

\textbf{SumQE (\texttt{SQE})} \cite{xenouleas-etal-2019-sum}  predicts five linguistic quality scores via multi-head regression on human scores. We use the coherence head of the model trained on all three DUC datasets.\footnote{\url{https://archive.org/download/sum-qe/BERT_DUC_all_Q5_Multi\%20Task-5.h5}}

\textbf{GRUEN (\texttt{GRN})} \cite{zhu-bhat-2020-gruen} is a collection of unsupervised measures for linguistic quality that combines multiple unsupervised heuristics.

\textbf{BARTScore (\texttt{BAS})} \cite{yuan_bartscore_2021} uses the probability of a summary under a pre-trained BART model as a score. We use the variant fine-tuned on CNN/DM summaries in the source-to-summary configuration, as suggested by the authors.

Finally, we include an upper and lower bound:
\texttt{RND} assigns each summary a uniformly chosen score between 0 and 1. For \texttt{HUM}, we use the Summ\-Eval
human annotations and select the annotator with the worst overall correlation to the remaining annotators and use their scores as predictions.%
\footnote{We note that unlike automatic measures, humans may only differentiate among five classes. We might thus underestimate actual human performance.}

We train all shuffling models on the WSJ corpus of newswire articles, which is frequently used in coherence modelling \cite{elsner-charniak-2011-extending, guinaudeau-strube-2013-graph, moon-etal-2019-unified, mohiuddin_rethinking_2021}. We also train models using the same technique on reference summaries from the train portion of CNN/DM. For \texttt{EEG} we also estimate model parameters on both datasets.
For WSJ, we follow the original implementations regarding the number of shuffled samples. For CNN/DM, we only use a single shuffled instance per summary, as it is larger by two orders of magnitude (WSJ: 1,400; CNN/DM: 287,113 documents before shuffling). Detailed accounts of our experiments with each CM are found in Appendix \ref{app:implementations}.

\section{Results}

\begin{table*}[!ht]
    \centering
    \small\begin{tabular}{|l|r|r|r|r|r|r|}
    \hline
    Metric & $\tau_{intra}$ & $\tau_{pair}$ & $\tau_{sum}$ & $\tau_{sys}$ & Acc.$_{pair}$ \\\hline
\hline
 HUM     &	+0.75 \tiny{(+0.70 +0.79)}  &    +0.81	\tiny{(0.76, 0.85)}			  &  +0.81 \tiny{(+0.77 +0.84)}   &  +0.91 \tiny{(+0.71 +1.00)}   &  +0.77 \tiny{(+0.71 +0.81)}	\\
RND     &	-0.00 \tiny{(-0.06 +0.05)}  &   -0.00 \tiny{(-0.07 +0.06)}  &   +0.00 \tiny{(-0.05 +0.05)}  &   +0.09 \tiny{(-0.41 +0.53)}  &   +0.50 \tiny{(+0.46 +0.54)}	\\\hline
EGR     &	-0.04 \tiny{(-0.12 +0.04)}  &     -0.11 \tiny{(-0.19, -0.02)}			  &  -0.09 \tiny{(-0.16 -0.01)}   &  -0.25 \tiny{(-0.59 +0.10)}   &  +0.40 \tiny{(+0.36 +0.44)}	\\
EEG C/D &	+0.02 \tiny{(-0.07 +0.10)}  &   +0.04 \tiny{(-0.10 +0.18)}  &   +0.06 \tiny{(-0.06 +0.17)}  &   -0.19 \tiny{(-0.68 +0.26)}  &   +0.52 \tiny{(+0.45 +0.59)}	\\
EEG WSJ &	+0.02 \tiny{(-0.06 +0.10)}  &   +0.00 \tiny{(-0.09 +0.11)}  &   +0.03 \tiny{(-0.06 +0.11)}  &   -0.19 \tiny{(-0.60 +0.26)}  &   +0.50 \tiny{(+0.44 +0.55)}	\\\hline
NEG C/D &	-0.07 \tiny{(-0.14 -0.00)}  &   -0.05 \tiny{(-0.14 +0.07)}  &   -0.06 \tiny{(-0.15 +0.03)}  &   -0.15 \tiny{(-0.61 +0.32)}  &   +0.47 \tiny{(+0.42 +0.53)}	\\
NEG DUC &	-0.08 \tiny{(-0.16 +0.01)}  &   -0.06 \tiny{(-0.18 +0.06)}  &   -0.07 \tiny{(-0.17 +0.04)}  &   -0.06 \tiny{(-0.49 +0.31)}  &   +0.47 \tiny{(+0.40 +0.53)}	\\
NEG WSJ &	-0.02 \tiny{(-0.08 +0.05)}  &   -0.08 \tiny{(-0.17 +0.00)}  &   -0.07 \tiny{(-0.15 +0.02)}  &   -0.43 \tiny{(-0.69 -0.05)}  &   +0.45 \tiny{(+0.41 +0.50)}	\\\hline
UNF C/D &	+0.04 \tiny{(-0.03 +0.11)}  &   +0.05 \tiny{(-0.05 +0.14)}  &   +0.06 \tiny{(-0.01 +0.13)}  &   +0.13 \tiny{(-0.33 +0.59)}  &   +0.53 \tiny{(+0.48 +0.57)}	\\
UNF WSJ &	+0.02 \tiny{(-0.05 +0.09)}  &   -0.11 \tiny{(-0.26 +0.03)}  &   -0.04 \tiny{(-0.15 +0.05)}  &   -0.09 \tiny{(-0.51 +0.39)}  &   +0.44 \tiny{(+0.36 +0.52)}	\\\hline
GRA DUC &	-0.04 \tiny{(-0.12 +0.03)}  &   -0.05 \tiny{(-0.16 +0.03)}  &   -0.06 \tiny{(-0.13 +0.01)}  &   -0.19 \tiny{(-0.65 +0.25)}  &   +0.47 \tiny{(+0.43 +0.52)}	\\
GRA C/D &	+0.08 \tiny{(+0.02 +0.15)}  &   +0.09 \tiny{(-0.02 +0.19)}  &   +0.11 \tiny{(+0.01 +0.18)}  &   +0.37 \tiny{(-0.07 +0.69)}  &   +0.55 \tiny{(+0.49 +0.60)}	\\
GRA WSJ &	+0.08 \tiny{(+0.01 +0.15)}  &   -0.01 \tiny{(-0.11 +0.10)}  &   +0.02 \tiny{(-0.06 +0.12)}  &   -0.09 \tiny{(-0.47 +0.37)}  &   +0.49 \tiny{(+0.44 +0.55)}	\\\hline
CCL C/D &	\textbf{+0.26} \tiny{(+0.19 +0.33)}  &   \textbf{+0.40} \tiny{(+0.31 +0.49)}  &   \textbf{+0.39} \tiny{(+0.31 +0.44)}  &   +0.62 \tiny{(+0.30 +0.86)}  &   \textbf{+0.71} \tiny{(+0.66 +0.76)}	\\
CCL WSJ &	+0.20 \tiny{(+0.12 +0.26)}  &   +0.35 \tiny{(+0.25 +0.46)}  &   +0.33 \tiny{(+0.24 +0.41)}  &   \textbf{+0.74} \tiny{(+0.40 +0.92)}  &   +0.69 \tiny{(+0.63 +0.74)}	\\\hline
BAS     &	+0.17 \tiny{(+0.08 +0.26)}  &   +0.37 \tiny{(+0.23 +0.51)}  &   +0.32 \tiny{(+0.20 +0.42)}  &   +0.72 \tiny{(+0.42 +0.89)}  &   +0.69 \tiny{(+0.62 +0.77)}	\\
GRN     &	+0.18 \tiny{(+0.12 +0.25)}  &   +0.26 \tiny{(+0.17 +0.35)}  &   +0.27 \tiny{(+0.19 +0.34)}  &   +0.72 \tiny{(+0.38 +0.89)}  &   +0.63 \tiny{(+0.58 +0.69)}	\\
SQE     &	+0.19 \tiny{(+0.13 +0.26)}  &   +0.26 \tiny{(+0.15 +0.36)}  &   +0.24 \tiny{(+0.15 +0.32)}  &   +0.51 \tiny{(+0.05 +0.80)}  &   +0.64 \tiny{(+0.58 +0.69)}	\\\hline

\end{tabular}
    \caption{Results on SummEval for all CMs. Correlation is expressed in Kendall's $\tau$. Numbers in brackets indicated 95\% CIs computed using bootstrap resampling \cite{deutsch-etal-2021-statistical} with 1000 samples. Highest are bold.}
    \label{tab:results}
\end{table*}

We present the correlation of all CMs with human coherence ratings in Table~\ref{tab:results}.
We report 
(average) Kendalls $\tau$ for all EMs introduced in Section~\ref{sec:em_intro}. For $C_{pair}$ we additionally report accuracy. Per-system scores for intra-system correlation can be found in Appendix~\ref{app:intra_sys}.

Focusing on $\tau_{sys}$ first, we find that \texttt{CCL}, \texttt{BAS}, \texttt{GRN} and to a lesser extent \texttt{SQE} achieve relatively high scores while the remaining CMs fail to outperform even the random baseline. However, inspection of $\tau_{sum}$, $\tau_{pair}$/$Acc_{pair}$ and $\tau_{intra}$ reveals that even these apparently strong CMs struggle to reliably assess coherence of individual summaries, with $\tau_{intra}$ being the most challenging regime.
Comparing CMs, \texttt{CCL C/D} is most promising across all EMs except $\tau_{sys}$, where scores are near indistinguishable due to high uncertainty.
Interestingly, we find that its advantage is greatest on $\tau_{intra}$, where its competitors exhibit particular weakness compared to other EMs.
These sharp score drops might suggest other EMs reflect some system-level confounders. In combination with the observation that confounder scores as reported in Table~\ref{tab:shortcuts} fall within the $95\%$ CI of most CMs on all EMs except $\tau_{intra}$ this prompts us to investigate CMs for potential biases in the following section.

\subsection{Detecting Biases of CMs} 

\begin{figure*}
    \centering
    \includegraphics[width=0.9\textwidth]{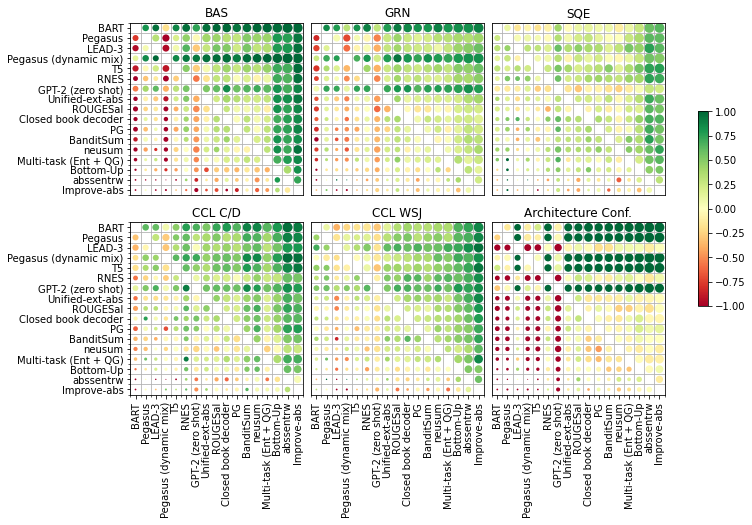}
    \caption{Bias Matrices for the best CMs. We also show the bias matrix for the architecture confounder for reference. See Figure~\ref{fig:bias_tutorial} for a brief tutorial to bias matrix analysis.}
    \label{fig:sys_level}
\end{figure*}

We have shown in Section~\ref{sec:confounders} that CMs can appear to correlate with human coherence judgements by exploiting system-level confounders. However, it is unclear to which extent this just holds for our artificial confounders or is also an issue in realistic CM evaluation. We therefore introduce \textit{bias matrices}, a tool that allows us to easily inspect the decisions made by a CM by separately analyzing \textit{consistent} and \textit{inverted} pairs of summaries from different summarizers.
\textit{Based on human scores}, we call a summary pair \textit{consistent} if the higher-scoring summary is produced by the summarizer with the higher average score, whereas we call a pair \textit{inverted} if the overall worse summarizer produces a stronger summary. We are specifically interested in finding instances where a CM ranks consistent pairs for a strong summarizer correctly, but fails to correctly rank its inverted pairs. This is indicative of a CM having a bias towards outputs of this particular summarizer, instead of measuring coherence. Since for strong systems, most pairs are consistent, this can still result in many correct comparisons.  %

Given predicted and human scores $P, H$ as in Section \ref{sec:em_intro} and systems $s_1, s_2$ with $s_1$ having a higher average human score than $s_2$, we define two new metrics. $\tau^+$ indicates the ability of a CM to rank consistent pairs, whereas $\tau^-$ indicates the same for inconsistent pairs. For $\tau^+$ we define:

\begin{align*}
    H^+ := \{ (d_i, d_j) | H_{(d_i,s_1)} > H_{(d_j,s_2)}\} \\
    P^+ := \{ (d_i, d_j) | P_{(d_i,s_1)} > P_{(d_j,s_2)}\} \\
    \tau^+ := \frac{2 |H^+ \cap P^+| - |H^+|}{|H^+|}
\end{align*}

For $\tau^-$ we invert the comparisons.\footnote{If $s_1$ is better than $s_2$ on every document, $\tau^-$ is undefined. In this case, biased and unbiased CMs are indistinguishable.}
Both $\tau^+$ and $\tau^-$  are bounded between -1 and 1. If the ranking is -1, this indicates the ranking is always incorrect, 1 always correct. %
To derive the $|S| \times |S|$ bias matrix $\mathrm{T}$, we order systems $s_1 \ldots s_n$ in descending order of their average human score. We then have:
\begin{equation*}
    \mathrm{T}_{ij} := \begin{cases}
    \tau^+(s_i, s_j) & i < j \\
    \tau^-(s_j, s_i) & i > j \\
    0 & i = j
    \end{cases}
\end{equation*}

\begin{figure*}
    \centering
    \includegraphics[width=\textwidth]{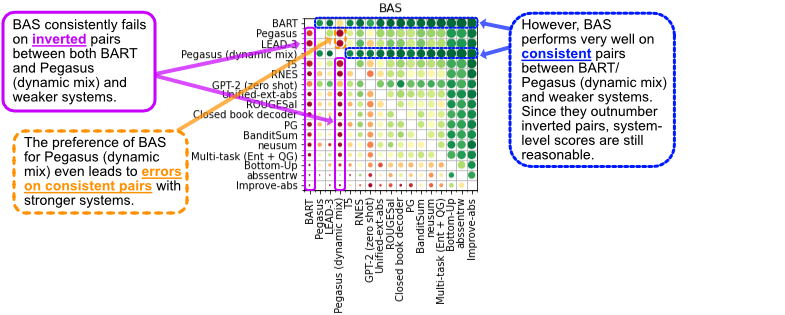}
    \caption{Bias matrix for \texttt{BAS} with specific analysis for BART and Pegasus. The upper triangular matrix indicates $\tau^+$ for the given summarizer pair, the lower $\tau^-$. The area of each circle is proportional to the number of pairs in $H+$/$H^-$ for the cell. To read off the behaviour of the CM on a specific summarizer, we follow both the corresponding row and column. A high score in the row, combined with a low score in the corresponding cell in the column implies the CM is biased towards generations by this particular summarizer.}
    \label{fig:bias_tutorial}
\end{figure*}

We visualize $\mathrm{T}$ for the most promising CMs in Figure~\ref{fig:sys_level}. To aid interpretation, we provide an annotated version for scores generated by \texttt{BAS} in Figure~\ref{fig:bias_tutorial}. We find that \texttt{GRN} and \texttt{BAS} show a very strong preference for summaries generated by BART, ranking them almost universally higher even when this disagrees with human judgements. In case of \texttt{BAS} this is unsurprising, since BART and \texttt{BAS} use the same underlying model. For \texttt{GRN} the reason is less clear, though analysis in Section~\ref{sec:gruen} suggests that it might rely on the higher grammaticality of BART output.
For the other CMs, biases are less evident, though \texttt{CCL} \texttt{C/D} shows a slight preference for BART and Pegasus and \texttt{CCL} \texttt{WSJ} has a slight bias towards LEAD and GPT-2.

\section{CM Analysis}

\subsection{Correlation with Shuffle-Performance} \label{sec:shuffle}

\citet{mohiuddin_rethinking_2021} have shown that the  performance of CMs on the shuffle task is not predictive for performance on summary coherence evaluation. However, at the same time, the shuffling-based \texttt{CCL} shows comparatively strong performance in our experiments.
To better understand the relation between shuffling and summary coherence, we test the ability of all CMs to discriminate shuffled and non-shuffled \textit{reference summaries} from the test split of CNN/DM. Results are in 
Table~\ref{tab:shuffle}.

Of the CMs that perform best on coherence evaluation (see Table~\ref{tab:results}), most also perform well on the shuffling task (\texttt{CCL}, \texttt{BAS}, \texttt{SQE}). Only \texttt{GRN} fails on this task. This is troubling as we would expect any CM that is able to identify coherent summaries on SummEval to be able to identify at least some shuffled reference summaries. This suggests that \texttt{GRN} models coherence only indirectly via proxy variables, which we elaborate on in Section~\ref{sec:gruen}.

For the entity-based measures \texttt{EGR}, \texttt{EEG} and \texttt{NEG}, their difficulties on  the SummEval dataset are also reflected in the shuffle task. This suggests that these CMs struggle generally on CNN/DM-style summaries. In Section~\ref{sec:entities} we demonstrate that this is due to the overall lack of entity overlap in this dataset. Finally, \texttt{UNF} \texttt{C/D} and \texttt{GRA} are outliers in that they show shuffle performance on CNN/DM that is similar or better than \texttt{SQE} but still perform near random on Summ\-Eval coherence modelling. We investigate this in Section~\ref{sec:length}.

\begin{table*}[ht!]
    \centering
\small\begin{tabular}{|l|c|c|c|c|c|c|c|c|c|}
\hline
Corpus  &       EGR     &       EEG     &       NEG     &       GRA     &       UNF     &       CCL     &       BAS     &       GRN   &SQE\\\hline
C/D     &       0.426   &       \begin{tabular}[x]{@{}c@{}}$0.523_{(c)}$\\$0.498_{(w)}$\end{tabular}    &       \begin{tabular}[x]{@{}c@{}}$0.524_{(c)}$\\$0.603_{(w)}$\\$0.522_{(d)}$\end{tabular}    &       \begin{tabular}[x]{@{}c@{}}$0.838_{(c)}$\\$0.623_{(w)}$\\$0.439_{(d)}$\end{tabular}    &       \begin{tabular}[x]{@{}c@{}}$0.803_{(c)}$\\$0.589_{(w)}$\end{tabular}    &       \begin{tabular}[x]{@{}c@{}}$0.929_{(c)}$\\$0.862_{(w)}$\end{tabular}   &       0.896   &       0.504   &       0.707\\\hline\hline

WSJ (orig.)     &       0.889       &       0.840    &       0.855    &       0.924   &       0.93    &       0.97    &       -       &       -     &- \\
\hline
\end{tabular}
\caption{Shuffle accuracies on CNN/DM for 1000 randomly sampled reference summaries. (c) means that the model was trained on CNN/DM   (w) on WSJ and (d) on DUC03. Baseline accuracy would be 50\%. For reference, we also list originally reported shuffle results on full WSJ articles as originally reported where applicable.}
    \label{tab:shuffle}
\end{table*}

\subsection{GRUEN} \label{sec:gruen}

\begin{table}
    \centering
    \begin{tabular}{|l|c|c|c|c|}
    \hline
        &   Cola    &  Redun.   &   LM      & Focus  \\\hline
Cola    &   0.57    &  \textbf{0.71}     &   0.59    &   0.63 \\
Redun.  &           &  0.51     &   0.57    &   0.51 \\
LM      &           &           &   0.15    &   0.35 \\
Focus   &           &           &           &   0.49 \\\hline
    \end{tabular}
    \caption{Performance of \texttt{GRN} constituent measures. Cells indicate system-level correlation of the combination of the respective measures. Individual measure performance is indicated on the diagonal.}
    \label{tab:gruen}
\end{table}

\texttt{GRN} works well for system-level correlation yet is incapable of solving the shuffle task. This prompts us to investigate its individual components. In the reference implementation, \texttt{GRN} computes the sum of three scores to determine the overall score.\footnote{The coherence score reported in the paper is not part of the reference implementation. We have confirmed that this is intentional in personal communication with the authors.}
\textbf{Grammaticality} is assessed per sentence by a classifier trained on the 
CoLA corpus \cite{warstadt-etal-2019-neural} and the average log probability under a BERT model.
\textbf{Redundancy} is estimated by a fixed penalty whenever any sentence pair has token overlap above a predetermined threshold. \textbf{Focus} is scored by word-mover-similarity \cite{kusner_word_2015} of neighbouring sentences.

Table~\ref{tab:gruen} shows the system-level correlation of the individual scores and all pairwise combinations. CoLA plus redundancy alone account for almost the full system-level correlation of $0.72$. %
Since neither score is dependent on sentence order, they can by design not fully account for summary coherence, raising considerable doubt
about the generalizability of \texttt{GRN}s  performance on this task.

\subsection{Entity Driven Measures} \label{sec:entities}

To explain why \texttt{EEG}, \texttt{EGR} and \texttt{NEG} perform poorly even on the shuffle task, we investigate the role of entity (re-)occurrences in CNN/DM summaries. Table~\ref{tab:entities} shows that both reference summaries and SummEval data have very little lexical entity overlap in between sentences.\footnote{
As determined by the Brown Coherence Toolkit. See Appendix~\ref{app:implementations}.
} A considerable number of summaries in both SummEval and CNN/DM show no entity overlap between any of their sentences. Therefore entity-based models are inherently limited, at least when using lexical overlap to determine
entity re-occurrence. We leave a thorough investigation of solutions like better coreference resolution  or using embedding based methods as in \citet{mesgar-strube-2016-lexical} to future work.

\begin{table}
    \centering
    \begin{tabular}{|l|c|c|}\hline
      Corpus        & Docs & Sents \\\hline
    CNN/DM Ref.     & 0.287 & 0.458 \\
    SummEval        & 0.178 & 0.301 \\
    DUC03           & 0.014 & 0.121 \\\hline
    \end{tabular}
    \caption{Proportion of documents without any entity overlap, as well as average ratio of sentences without entity links per document for various datasets.}
    \label{tab:entities}
\end{table}

\subsection{Global Training vs. Pairwise Ranking} \label{sec:length}

While CMs that fail the in-domain shuffling task are likely to be unsuitable for CNN/DM summaries, it is less clear why CMs with reasonable shuffle performance fail on  SummEval like \texttt{UNF} \texttt{C/D} and \texttt{GRA} \texttt{C/D}. We theorize that one reason is that both \texttt{UNF} and \texttt{GRA} are trained on a margin-based ranking loss between shuffled and non-shuffled variants of the \textit{same} document, which implies that both have the same tokens and number of sentences. 
The training loss thus does not impose constraints on the behaviour of the function between inputs of \textit{different} lengths and tokens. Since SummEval, unlike e.g. DUC, has no agreed upon length constraint, this is problematic.\footnote{Summarizer length statistics are in Appendix~\ref{app:length}.}
In contrast, the classification objective of \texttt{CCL} enforces a globally correct ranking of shuffled vs. unshuffled documents.

\begin{figure}
    \centering
    \includegraphics[width=0.45\textwidth]{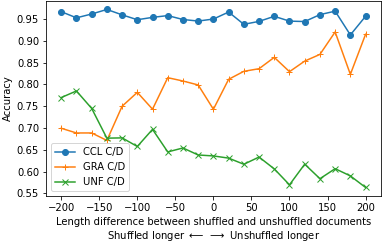}
    \caption{Ranking accuracy between shuffled and origi-nal summaries of different lengths (in characters). We sample 10,000 pairs and group them in buckets of 20 characters and clamp differences between -200 and 200.}
    \label{fig:length}
\end{figure}

Verifying this hypothesis on SummEval directly is difficult, since summary length is deeply confounded with the generating summarizer.
However, we can  investigate the ability of CMs to correctly rank documents of different lengths and content by modifying the shuffle test to compare reference summaries to shuffled variants of \textit{different} reference summaries. Figure~\ref{fig:length} shows the relation between the difference in length between the shuffled and unshuffled summaries and the ranking accuracy of the CMs.
 \texttt{UNF} performs very poorly on the task, especially if the original summary is long. \texttt{GRA}, on the other hand, prefers longer documents, even if they are shuffled. In contrast, \texttt{CCL} is consistently able to correctly rank summaries regardless of length difference.
Thus, for both \texttt{UNF} and \texttt{GRA} comparing documents of different lengths and content is a major obstacle. The stability of \texttt{CCL} suggests that replacing pairwise ranking with a classification objective is a direct fix to this issue. These results are also consistent with parallel work by \citet{jwalapuram-etal-2022-rethinking} who extend the pairwise shuffle-task to consider multiple negative examples. They find that including negative samples from different documents in the negative set during training improves model performance on downstream tasks.

\section{Conclusion}

We have investigated the performance of a wide array of CMs for summary evaluation that have not been previously systematically compared.
Our investigations show that CMs must be carefully evaluated in order to avoid rewarding the modelling of shallow, system-level confounders, that are unlikely to generalize.
We thus recommend researchers report our newly suggested intra-system correlation \textit{alongside} other EMs and use bias matrices to understand unexpected drops when going from system-level to   intra-system correlation.

There is considerable need to improve CMs before they become practical for summary coherence modelling. Our results point towards the following lessons for future work. Firstly, 
CNN/DM summaries are not amenable to entity-based analysis without considerable additional work to improve entity detection. Secondly, self-supervised training via the shuffle task shows the greatest promise for future improvements. However, we show that good shuffle performance does not naturally transfer to coherence evaluation for settings with documents of different lengths and contents. Training in a classification setup instead of the more common pairwise setup provides an effective fix for this.

\bibliography{anthology,references}
\bibliographystyle{acl_natbib}

\appendix

\section{Implementation Details} \label{app:implementations}

\subsection{Extended Entity Grid \texttt{(EEG)}}

We use the original implementation that is part of the Brown Coherence Toolkit\footnote{\url{https://web.archive.org/web/20200505174052/https://bitbucket.org/melsner/browncoherence}}.
For preprocessing, we use the Stanford parser\footnote{\url{https://nlp.stanford.edu/software/lex-parser.shtml}}. We identify entities using OpenNLP as suggested in the README.

For WSJ we used the pretrained \texttt{f-wsj} model provided in the toolkit. For CNN/DM we trained our own model. We found that the implementation ran out of memory on the 287,011 instances in CNN/DM on our machine with 32GB of RAM. We thus limited the instances considered for CNN/DM to 10\% of the original dataset (28,701).

\subsection{Entity Graph \texttt{(EGR)}}

Since there is no reference implementation of the Entity Graph, we implement our own version based on the grid created by the Brown Coherence Toolkit. We use the $P_{Acc}$ measure with distance penalty which performed best in the original paper.

\subsection{Neural Entity Grid \texttt{(NEG)}}

Since no models are publicly available, we train new models for all settings using the reference implementation\footnote{\url{https://github.com/datienguyen/cnn_coherence}}.

For DUC03 and WSJ we use the entity grids and training pairs  provided by the authors in the repository. These were also created using the Brown Coherence Toolkit. For CNN/DM we create our own examples, following the original settings. We found that the original implementation of the shuffling procedure leaves artifacts in the data since the row order is unchanged between shuffled and unshuffled documents. However, for unshuffled documents the order of rows in the entity grid roughly corresponds to the order of entities in the sentences, whereas for shuffled documents this is not the case. Since this can be picked up by the convolutional network for short documents, we modify the input data to randomly shuffle row order for each instance.

For the shuffling tasks on WSJ we use the reported hyperparameters, which we also use for CNN/DM.
For DUC, no hyperparameters were reported, so we use the built-in hyperparameter search. We achieve the best results using the parameters reported in Table \ref{tab:neural_grid}.

\begin{table}
    \centering
    \begin{tabular}{|l|c|}
    \hline
        Embedding Size & 100 \\
        Batch Size & 64 \\
        Pool Length & 6 \\
        Window Size & 6 \\
        Number of Filters & 150 \\
        Hidden Size & 250 \\\hline
    \end{tabular}
    \caption{Best hyperparameters for the neural entity grid on DUC03.}
    \label{tab:neural_grid}
\end{table}

\subsection{Graph-based Model (\texttt{GRA})}

We use the original implementation.\footnote{\url{https://github.com/UKPLab/emnlp2021-neural-graph-based-coherence-model}}
For WSJ, we use the provided pretrained model. For DUC and CNN/DM we train the model using default settings, which includes an ELMo embedding layer. The graph-representation is created from an entity grid representation as provided by the Brown Coherence Toolkit.

\subsection{Unified Coherence Model \texttt{(UNF)}}

We use the original implementation.\footnote{\url{https://github.com/taasnim/unified-coherence-model}} We train new models for CNN/DM and WSJ using default settings. In the original implementation, scores are computed using a sum over coherence scores for windows of three sentences each, since in their pairwise evaluation, samples always have the same length. In our experiments, we use the mean over the windows instead to normalize for length. For completeness, we also conducted experiments using the original setting, which did not lead to any improvement.

\subsection{Coherence Classifier \texttt{(CCL)}}

We originally experimented with the pretrained WSJ model provided by the authors of \citep{laban-etal-2021-transformer}.\footnote{\url{https://github.com/tingofurro/shuffle_test}}
However, we found that the model achieved near-random scores when evaluated on SummEval for reasons that are difficult to ascertain as the original training code is unavailable.
We thus train our own coherence classifier models for both CNN/DM and WSJ.
We use the \texttt{roberta-large} model as implemented in the huggingface library \citep{wolf-etal-2020-transformers} in a sequence classification setup. We use a learning rate of $2e-6$ and train for a maximum of six epochs. We select the best model using f1-score on the validation set.

\subsection{BARTScore \texttt{(BAS)}}

We reimplement the finetuned BARTScore variant using the \texttt{bart-large-cnn} checkpoint from the huggingface library. Since the original model is evaluated using Spearman's $\rho$, we separately verified that it exactly reproduces the reported results.

\subsection{GRUEN \texttt{(GRN)}}

We use the scores provided by the official reference implementation.\footnote{\url{https://github.com/WanzhengZhu/GRUEN}}

\subsection{SumQE \texttt{(SQE)}}

We use the scores provided by the official reference implementation.\footnote{\url{https://github.com/nlpaueb/SumQE}} We use the Q5 head of the model jointly trained on all three DUC datasets.\footnote{\url{https://archive.org/download/sum-qe/BERT_DUC_all_Q5_Multi\%20Task-5.h5}}

\subsection{Hardware}

All experiments that include neural network training (i.e. \texttt{NEG}, \texttt{GRA}, \texttt{UNF}, \texttt{CCL}) were run on a single node with four Quadro RTX 6000 GPUs.

\section{Detailed Intra-System Correlation Results} \label{app:intra_sys}

\begin{figure*}
    \centering
    \includegraphics[width=1\textwidth]{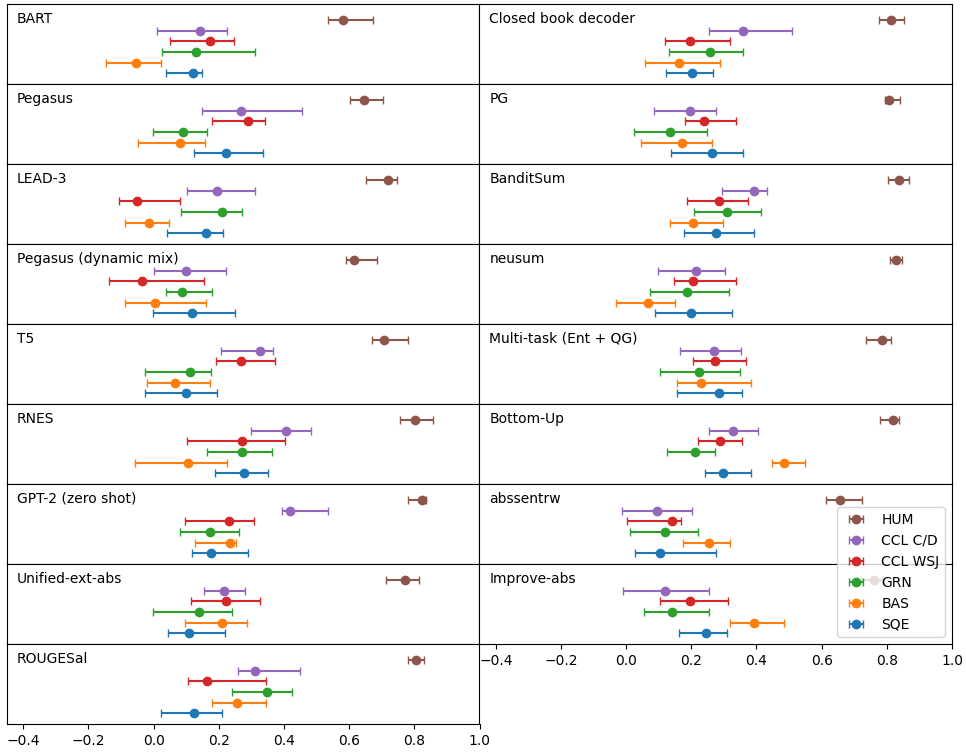}
    \caption{Intra-system correlations of the best CMs as well as the human upper bound on the SummEval dataset. Bars indicate 95\% confidence intervals determined by bootstrap resampling with 1000 samples.}
    \label{fig:intra_sys_details}
\end{figure*}

Figure \ref{fig:intra_sys_details} shows the individual intra-system correlations for all summarizers in SummEval for the best CMs and the human upper bound.
We find that CMs struggle across the whole range of summarizers, including summarizers with high variance in coherence scores, where we would expect the task to be easier.
Furthermore, we find none of the available CMs can consistently outperform all others. For example, 
\texttt{BAS} outperforms other CMs on Bottom-Up and Improve-Abs, but performs significantly worse on the top systems, including BART itself.

\section{Length Statistics} \label{app:length}

\begin{figure*}
    \centering
    \includegraphics[height=0.8\textheight]{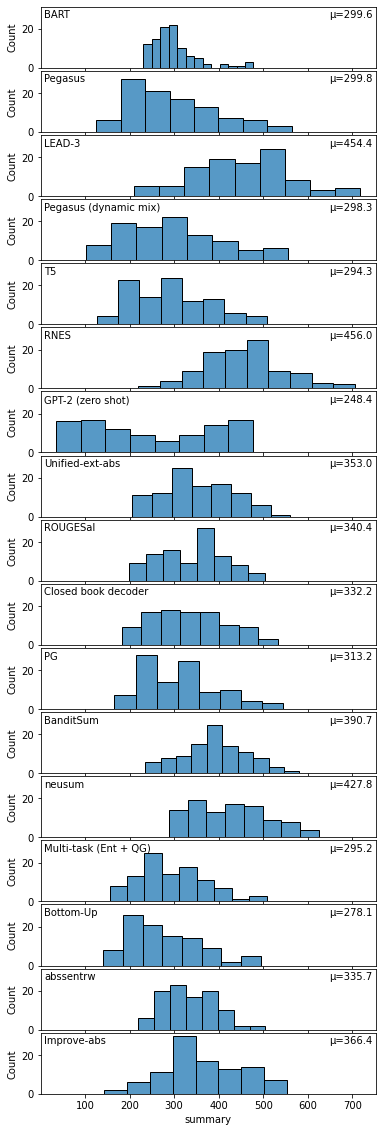}
    \caption{Histograms of the lengths of summaries generated by the summarizers in SummEval and their mean lengths. Both in characters.}
    \label{fig:lengths}
\end{figure*}

We present the length distribution of summarizer outputs on SummEval in Figure \ref{fig:lengths}.

\end{document}